%% file: main.tex
\documentclass[letterpaper, 10 pt, conference]{ieeeconf}  
\usepackage{amssymb}
\usepackage{afterpage}
\IEEEoverridecommandlockouts                              

\input{macros}

\title{LiDAR-EDIT: LiDAR Data Generation by Editing the Object Layouts in Real-World Scenes}
    
\begin{document}
\author{Shing-Hei Ho$^{1}$, Bao Thach$^{1}$, and Minghan Zhu$^{2,3}$%
\thanks{$^{1}$University of Utah, Salt Lake City, UT 84112, USA.
        {\tt\small \{shinghei.ho, bao.thach\}@utah.edu}}%
% \thanks{$^{1}$Robotics Center and Kahlert School of Computing, University of Utah, Salt Lake City, UT 84112, USA.
%         {\tt\small \{shinghei.ho, bao.thach\}@utah.edu}}%
\thanks{$^{2}$University of Michigan, Ann Arbor, MI 48109, USA.
        {\tt\small minghanz@umich.edu}}%
\thanks{$^{3}$University of Pennsylvania, Philadelphia, PA 19104, USA.
        {\tt\small minghz@seas.upenn.edu}}%
}
\maketitle

\urlstyle{same}

\begin{abstract}
% We propose a framework to edit real-world LiDAR scans with novel object layouts while preserving a realistic background environment. Compared to the synthetic data generation frameworks where LiDAR point clouds are generated from scratch, our framework focuses on new scenario generation in a given background environment, and our method also provides object labels for the generated data. This approach ensures the generated data remains relevant to the specific environment, aiding both the development and the evaluation of algorithms in real-world scenarios. Compared with novel view synthesis, our framework allows the creation of counterfactual scenarios with significant changes in the object layout and does not rely on multi-frame optimization. 
\bao{We present $\lidaredit$, a novel paradigm for generating synthetic LiDAR data for autonomous driving. Our framework edits real-world LiDAR scans by introducing new object layouts while preserving the realism of the background environment.
Compared to end-to-end frameworks that generate LiDAR point clouds from scratch, $\lidaredit$ offers users full control over the object layout, including the number, type, and pose of objects, while keeping most of the original real-world background. 
Our method also provides object labels for the generated data.
Compared to novel view synthesis techniques, our framework allows for the creation of counterfactual scenarios with object layouts significantly different from the original real-world scene. 
$\lidaredit$ uses spherical voxelization to enforce correct LiDAR projective geometry in the generated point clouds by construction. During object removal and insertion, generative models are employed to fill the unseen background and object parts that were occluded in the original real LiDAR scans. 
% Our object removal module leverages generative background inpainting, while object insertion is achieved through point cloud completion and spherical projection.
Experimental results demonstrate that our framework produces realistic LiDAR scans with practical value for downstream tasks. 
% This represents a significant advancement and a crucial stepping stone for the development of LiDAR-based autonomous driving systems. 
}
% In our framework, the object removal and insertion are supported by generative background inpainting and object point cloud completion, and the entire pipeline is built upon spherical voxelization, which realizes the correct LiDAR projective geometry by construction. Experiments show that our framework generates realistic LiDAR scans with object layout changes and benefits the development of LiDAR-based self-driving systems. 
% We published all code, data, and visualization associated with this paper at 
Project website with open-sourced code: \url{https://sites.google.com/view/lidar-edit}
\end{abstract}
% \vspace{-5pt}

%%%%%%%%%%%%%%%%%%%%%%%%%%%%%%%%%%%%%%%%%%%%%%%%%%%%%%%%%%%%%%%%%%%%%%%%%%%%%%%%
\section{Introduction}\label{sec:intro}
\input{introduction}

\section{Related Work}\label{sec:related_work}
\input{related_works}

\section{Problem Formulation}\label{sec:problem}
\input{prob_formulation}

\section{Method}\label{sec:method}
\input{methods}

\section{Experiments and results}\label{sec:experiments}
\input{experiment}

\section{Conclusions}\label{sec:conclusions}
\input{conclusions}

% \section*{ACKNOWLEDGMENT}
% This work was supported in part by NSF Awards \#2024778 and \#2133027.

%%%%%%%%%%%%%%%%%%%%%%%%%%%%%%%%%%%%%%%%%%%%%%%%%%%%%%%%%%%%%%%%%%%%%%%%%%%%%%%%

\clearpage\newpage
\bibliographystyle{IEEEtran}
\bibliography{bibliography}

\end{document}

%% file: macros.tex
\usepackage{graphicx}
\usepackage{url}
\usepackage{multicol}
\usepackage{graphicx}
\usepackage{color}
\usepackage[pagebackref=true,breaklinks=true,colorlinks,bookmarks=false]{hyperref}
\usepackage[export]{adjustbox}
\usepackage{subcaption}
\usepackage{wrapfig}

\usepackage{multicol}
\usepackage{booktabs}
\usepackage{algorithm}
\usepackage[noend]{algpseudocode}
\usepackage{float}
\usepackage{amsmath, amssymb}
\usepackage{amsfonts}
\usepackage[skip=2pt,font=small,labelfont=bf]{caption}
\usepackage{bm}
\usepackage{epstopdf}
\usepackage[dvipsnames]{xcolor}

\usepackage{todonotes}

\definecolor{purple}{RGB}{210, 0, 210}
\definecolor{dgreen}{RGB}{34, 139, 32}

\definecolor{international_orange}{RGB}{240, 74, 0}

\newcommand{\baot}[1] {\textcolor{black}{#1}}

\newcommand{\bao}[1] {\textcolor{black}{#1}}

\usepackage{cite} % Group multiple citations into brackets

\newcommand{\static}{\mathcal{S}}
\newcommand{\dynamic}{\mathcal{D}}

\newcommand{\combinedscene}{\mathcal{O}}
\newcommand{\obsfunc}{f}
\newcommand{\pcloud}{\mathcal{P}}
\newcommand{\gtpcloud}{\pcloud_\mathrm{o}}
\newcommand{\syntheticpcloud}{\pcloud_T}

\newcommand{\ggn}{\emph{DefGoalNet}}
\newcommand{\DeformerNet}{\emph{DeformerNet}}
\newcommand{\lidaredit}{\text{LiDAR-EDIT}}

\newcommand{\mhz}[1] {\textcolor{black}{#1}} %alan switch this

%% file: introduction.tex
\begin{figure*}[t]
    \centering
    \includegraphics[width=\textwidth]{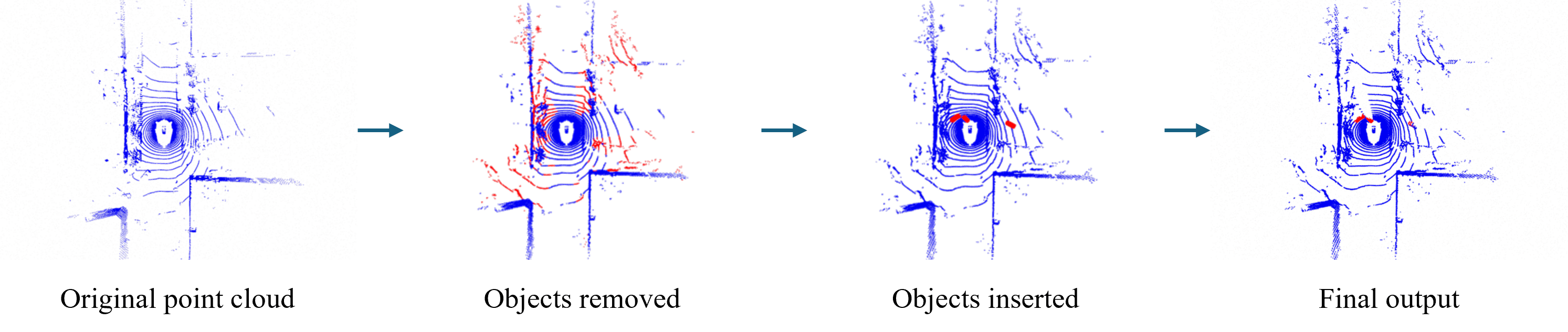}
    \caption{Overview of our novel LiDAR editing paradigm. Given the point cloud of a real-world LiDAR scan, we want to freely change the objects and their poses while preserving the background environment. This requires filling the background when objects are removed, and handling occlusion and LiDAR scan projection when new objects are inserted. Edited points are highlighted in red. }
    \label{fig:title}
    \vspace{-10pt}
\end{figure*}

\baot{Data plays a crucial role in the advancement of self-driving technology. High-quality data that accurately captures the complexity and diversity of real-world driving scenes is essential for both the development and testing of autonomous systems. However, real-world data collection is inherently limited due to its high cost and potential safety hazards, especially when capturing long-tail corner cases, which occur infrequently and involve significant risk.
Furthermore, real-world data simply represents a fixed recording of past events, making it inflexible for modifying specific factors to analyze their causal impact on self-driving algorithm performance. }

% \bao{Unfortunately, despite significant efforts to construct large-scale AV datasets [ref], often comprising thousands of hours of driving data, they remain insufficient [ref]. On one hand, data collection requires direct interaction with the physical world, which is both slow and extremely expensive. On the other hand, allowing AVs to explore risky, novel scenarios in real-world environments is highly undesirable, as it could endanger human lives or cause damage to the vehicle itself.}

These challenges have motivated research focused on generating synthetic data for autonomous driving. Early efforts rely on gaming engines and graphical rendering. This approach generates data from highly configurable virtual environments, but it induces a non-negligible sim-to-real gap, and the scalability is limited. 
Recent advancements in generative models \cite{ho2020denoising, radford2015unsupervised}, have enabled end-to-end data generation \cite{caccia2019deep, zyrianov2022lidargen}, making large-scale and realistic synthetic data possible. However, this approach comes with a significant drawback: \bao{users have limited control over the generated data, such as precisely specifying where a newly generated car will be positioned within the scene.}
% However, users generally have less control over the generated data. 
Another notable trend is novel-view synthesis from real-world sequential observations \cite{huang2023nfl}, allowing the rendering of a recorded scenario from different viewpoints. However, the variation flexibility of the rendered data is limited by the visible signal.

In this paper, we propose a new paradigm for synthetic data generation: generative 3D scene editing with controllable novel object layouts. We focus on LiDAR scans—a widely used and reliable sensory modality in autonomous vehicles. This paradigm strikes a balance between three key factors: \textit{faithfulness}, \textit{controllability}, and \textit{scalability}. \textit{Faithfulness}: the generated data are \baot{only partial} modifications of the real-world data, preserving the characteristics of the real environment and objects in the original data, thereby reducing the \baot{synthetic-to-real} domain gap. 
% \textit{Controllability}: the generated data follows the user's specification of the object layout, including the amount, type, and pose of the objects, allowing convenient comparative analysis. 
\textit{Controllability}: users have full control over the object layout in the generated data, including the number, type, and pose of objects, which better facilitates comparative analysis.
\textit{Scalability}: our paradigm does not rely on a physics engine, prebuilt 3D assets, or reconstruction of the whole scene. Instead, it is generative, facilitating efficient and scalable data creation based on vast amounts of existing real-world data.

An illustration of our proposed method for this paradigm, $\lidaredit$, is shown in Fig.~\ref{fig:title}. Our framework first removes unwanted objects from the scene and inpaints missing regions using a generative model. We then insert the point clouds of objects, which are completed from their partial LiDAR scan, at user-specified poses. We use spherical voxelization to ensure that the occlusion and the scanning pattern follow the LiDAR projective geometry. Experimental results demonstrate that our framework enables flexible, controllable, and realistic LiDAR scene editing, offering significant benefits for both the development and testing of autonomous driving algorithms.
% \bao{}
% \bao{}

%% file: related_works.tex
% Recent progress in generative models has benefited the research of data generation for self-driving.
% We will review the literature related to LiDAR data generation. Most work did not directly address the same problem as targeted in this paper, but a lot of them are relevant and prepared the techniques that are useful in our problem. 
\bao{We review existing paradigms for LiDAR data generation in the literature.}

\subsection{LiDAR simulation}
% LiDAR simulation refers to the approach of building a 3D environment with 3D object assets and casting rays on the 3D environment to create point clouds through ray-surface intersection \cite{yue2018lidar}. While graphics-based simulation engines \cite{unrealengine, juliani2018unity, dosovitskiy2017carla} facilitate the rendering, the requirement of pre-built 3D assets limits the scalability, and the rendering and the quality of the 3D assets both contribute to the domain gap compared with real data. \cite{manivasagam2020lidarsim} constructed the 3D assets from real-world scans to avoid manual asset creation and used learning-based postprocessing to reduce the domain gap. Later work \cite{manivasagam2023towards} further investigated various LiDAR phenomena that could improve the fidelity of LiDAR simulators. 
\bao{LiDAR simulation creates a 3D environment with object assets and generates point clouds via ray-surface intersection. \cite{yue2018lidar}. Graphics-based engines \cite{unrealengine, juliani2018unity, dosovitskiy2017carla} face scalability limits from pre-built 3D asset requirements, while asset quality and rendering contribute to domain gaps. \cite{manivasagam2020lidarsim} used real-world scans to generate 3D assets and applied learning-based postprocessing to reduce this gap, while \cite{manivasagam2023towards} explored LiDAR phenomena to enhance simulator fidelity.}

% % simulation
% \cite{manivasagam2023towards}: investigated the affect of different LiDAR phenomena on the domain gap of lidar simulation. 
% \cite{yue2018lidar}: create point clouds with accurate point-level labels from a computer game. 
% \cite{manivasagam2020lidarsim}: first create the assets from real data, and then compose them into a scene and simulate the sensor with physics and machine learning. Collect real data from multiple trajectories in the same area, remove moving objects, aggregate and align the data, and create a mesh surfel representation of the background. Use a graphic engine to render, with learned ray-drop. 

\subsection{LiDAR novel-view synthesis}
% The novel-view synthesis (NVS) of LiDAR point clouds and LiDAR simulation share the same conceptual procedure of 3D construction and rendering, but today's LiDAR NVS realizes these two steps in a unified learnable framework called neural rendering. The learned implicit geometry-appearance model and the differentiable rendering are both optimized towards reproducing more realistic synthesis, allowing a smaller domain gap and a more automatic workflow than traditional simulation \cite{tao2023lidarnerf}. \cite{huang2023nfl} modeled LiDAR effects like secondary returns and ray dropping in the neural rending framework. \cite{yang2023unisim, zhang2024nerflidar} also leveraged image sequences for additional information. \cite{zheng2024lidar4d, wu2024dynfl} explicitly the temporal dimension to account for the moving objects. \cite{tao2024alignmif} rendered both LiDAR scans and camera images from the neural field. These approaches realized the playback of a real-world scenario with changed viewpoints and object states, but the pose of the observer and the objects cannot deviate too much from reality, or the rendering quality will deteriorate. 

\baot{LiDAR novel-view synthesis (NVS) and LiDAR simulation follow the same 3D construction and rendering process, but modern LiDAR NVS integrates both steps into a unified learnable framework called neural rendering. This approach optimizes a learned implicit geometry-appearance model through differentiable rendering for more realistic synthesis, reducing the domain gap and enabling a more automated workflow than traditional simulation \cite{tao2023lidarnerf}. \cite{huang2023nfl} incorporated LiDAR effects like secondary returns and ray dropping, while \cite{yang2023unisim, zhang2024nerflidar} used image sequences for additional data. \cite{zheng2024lidar4d, wu2024dynfl} accounted for moving objects by modeling the temporal dimension. \cite{tao2024alignmif} rendered both LiDAR scans and camera images from the neural field. These methods enable real-world scenario playback with altered viewpoints and object states, though significant deviations in the observer or object poses degrade rendering quality.}

\subsection{End-to-end LiDAR data generation}
End-to-end approaches bypass the two-step procedure of 3D construction and rendering. Inspired by the success of text- and image-based generative models, researchers proposed to generate LiDAR scans through a fully-differentiable neural network. This approach produces data with promising diversity and realism but has limited controllability. \cite{zyrianov2022lidargen} applied a diffusion model on equirectangular range-view images, leveraging the progress in image diffusion models. \cite{ran2024lidardiffusion} also used range images but compressed the data into a latent space before diffusion. It also supported conditioning like semantic maps and RGB images. \cite{xiong2023ultralidar} applied a generative transformer framework MaskGIT \cite{chang2022maskgit} in BEV-voxelized point cloud generation and allowed basic editing by swapping the discrete latent codes. Recent work started to generate temporally consistent sequences of LiDAR scans, incorporating the idea of world modeling \cite{zhang2023copilot4d}. \cite{zyrianov2024lidardm} took traffic maps as input conditions, improving the controllability of LiDAR sequence generation. The end-to-end generated point clouds generally do not come with corresponding labels, limiting their value in practical development.

% % world models
% \cite{zhang2023copilot4d}: end-to-end prediction of a sequence of future lidar frames, using an extended and improved framework of UltraLidar. Nuscenes, KITTI, and Argoverse 2. CD, L1 Med, AbsRel, ...
% \cite{zyrianov2024lidardm}: end-to-end generation of a sequence of lidar frames based on traffic layout condition. KITTI-360 and Waymo. Unconditional single frame generation evaluated with MMD and JSD. Layout-aware generation evaluated with CenterPoint performance (transfer and augmentation). 
% % end-to-end point cloud generation:
% \cite{zyrianov2022lidargen}: end-to-end generation through a stochastic
% denoising process in the equirectangular view. Can be conditioned. nuscenes and kitti360. Point cloud, no label. 
% \cite{xiong2023ultralidar}: end-to-end generation using MaskGIT. Can insert and remove objects but cannot specify the pose. The occlusion and deocclusion are not handled explicitly. Can also do lidar densification (tested through detection performance) and scene completion using the same framework. Generation from scratch is evaluated using MMD and JSD. 
% \cite{ran2024lidardiffusion}: range-image diffusion in latent space. Can take conditions including semantic maps, camera images, and texts. Bounding box conditions do not have demos in paper. Perceptual metrics FRID, FSVD, FPVD. Statistical metrics JSD and MMD. nuScenes, Semantic kitti, KITTI-360. 

\subsection{LiDAR scan modification}
Generative methods can also be used to apply modifications on an existing point cloud. For example, \cite{yang2024tulip} used a transformer to upsample a sparse point cloud. \cite{nunes2024lidiff} applied scene completion on a single LiDAR scan to obtain a dense and complete scene. \cite{singh2024genmm} allows inserting the same object in an RGB image and in a LiDAR point cloud in a consistent manner. Overall, the editing flexibility is limited. 
% % modification
% \cite{yang2024tulip}: point cloud upsampling in range view. DurLAR and KITTI datasets. MAE, Chamfer distance, and volumetric occupancy IoU. 
% \cite{nunes2024lidiff}: single-frame scene completion using point-based diffusion. SemanticKITTI and KITTi-360. 
% \cite{singh2024genmm}: Object insertion to a video of images and lidar frames. BDD100K and Waymo Open Dataset. AbsRel and L2 error on lidar object insertion. 

The task in this paper, LiDAR scan editing with novel object layout, is related to but different from all the tasks discussed above. Our output is connected to a real scene, similar to the NVS task, but we do not rely on reconstruction, which allows more flexibility in object manipulation. The task is generative but does not build scenes from scratch, emphasizing the controllability of object layouts. %, and allows a more flexible scene edition than previous work. 

%% file: prob_formulation.tex
We address the problem of generating realistic synthetic LiDAR data for autonomous driving. Rather than generating data from scratch, we build upon real-world scenes to manipulate foreground objects while preserving the background environment.

The static background of a LiDAR scene, $\static$, consists of stationary elements like streets, trees, and buildings, excluding foreground dynamic objects of interest such as vehicles, represented by $\dynamic$. A typical driving scene $\combinedscene$ includes both: $\combinedscene = \static \cup \dynamic$.
Since sensing the entire $\combinedscene$ from a single viewpoint is infeasible, we define an observation function $\obsfunc$ that transforms the scene into a partial-view point cloud: $\obsfunc(\combinedscene) = \obsfunc(\static, \dynamic) = \pcloud$, representing the real sensor output. 

The task of LiDAR generative editing can be formally defined as follows. We assume that we have a dataset of real-world LiDAR scans $P=\{\pcloud_i\}$. Given a true LiDAR scan $\gtpcloud = \obsfunc(\static, \dynamic_\mathrm{o})$, the goal is to generate a synthetic point cloud $\syntheticpcloud = \obsfunc(\static, \dynamic_T)$, where $\dynamic_\mathrm{o}$ is the true foreground objects in the scan, and $\dynamic_T$ is a modified set of foreground objects. Here, $\dynamic_T$ may consist of the same objects as $\dynamic_\mathrm{o}$ but with different poses, or it may contain entirely different objects. We assume that all possible objects in $\dynamic_T$ are observed in some $\pcloud_i \in P$. 

We tackle this problem by decomposing it into three sub-problems: estimating $\obsfunc(\static)$, $\dynamic_T$, and $\obsfunc(\static, \dynamic_T)$, corresponding to object removal, object point cloud completion, and object insertion. They are introduced in Sec. \ref{sec:method}. 

% by modifying the dynamic objects in $\gtpcloud$. The initial set of vehicles in $\gtpcloud$ is denoted as $\dynamic_\mathrm{o}$. We transform these objects into a new set $\dynamic_T$, resulting in a new point cloud: $\syntheticpcloud = \obsfunc(\static, \transform(\dynamic_\mathrm{o})) = \obsfunc(\static, \dynamic_T)$. 

% In this context, $\transform$ refers to a family of three possible transformations that can be applied to the initial vehicles. First, when $\transform \in \mathcal{SE}(3)$, it defines the \textit{homogeneous transformation} applied to the vehicles in the scene. Second, $\transform$ can represent a \textit{removal} operation, deleting a subset of objects in $\dynamic_\mathrm{o}$ from the scene. Third, $\transform$ can also represent an \textit{insertion} operation, injecting new vehicles to the scene. A combination of any of these three operations can also be applied.

% It is important to note that, while these transformations may seem straightforward, they pose significant challenges when applied to LiDAR point clouds. This is due to the nature of occlusion in LiDAR scans: for points along the same ray direction, only the point nearest to the sensor is retained, while the rest are occluded. As a result, naively applying a homogeneous transformation to points belonging to a vehicle can lead to unrealistic self-occlusion, inter-vehicle occlusion, and background-vehicle occlusion. The same reasoning applies to the removal and insertion operations. These challenges will be discussed in more detail in Sec.~\ref{sec:scene_representation}.

%% file: methods.tex
\subsection{Overview}
\begin{figure}
    \centering
    \includegraphics[width=\linewidth]{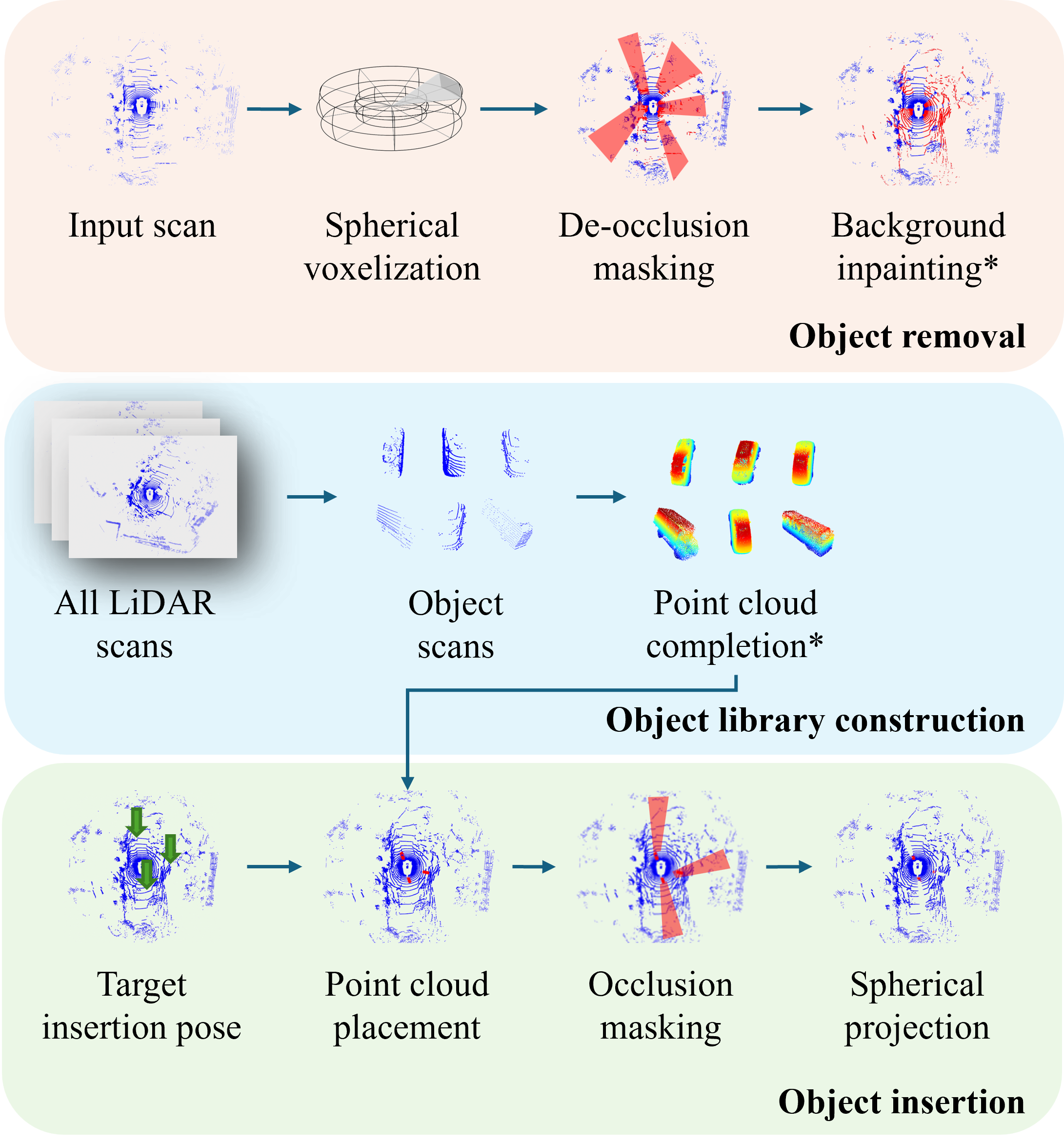}
    \caption{Overview of our novel $\lidaredit$ framework for LiDAR editing. Asterisk signs denote modules with generative models. }
    \label{fig:method_overview}
    % \vspace{-5pt}
\end{figure}

\begin{figure}
    \centering
    \includegraphics[width=0.9\linewidth]{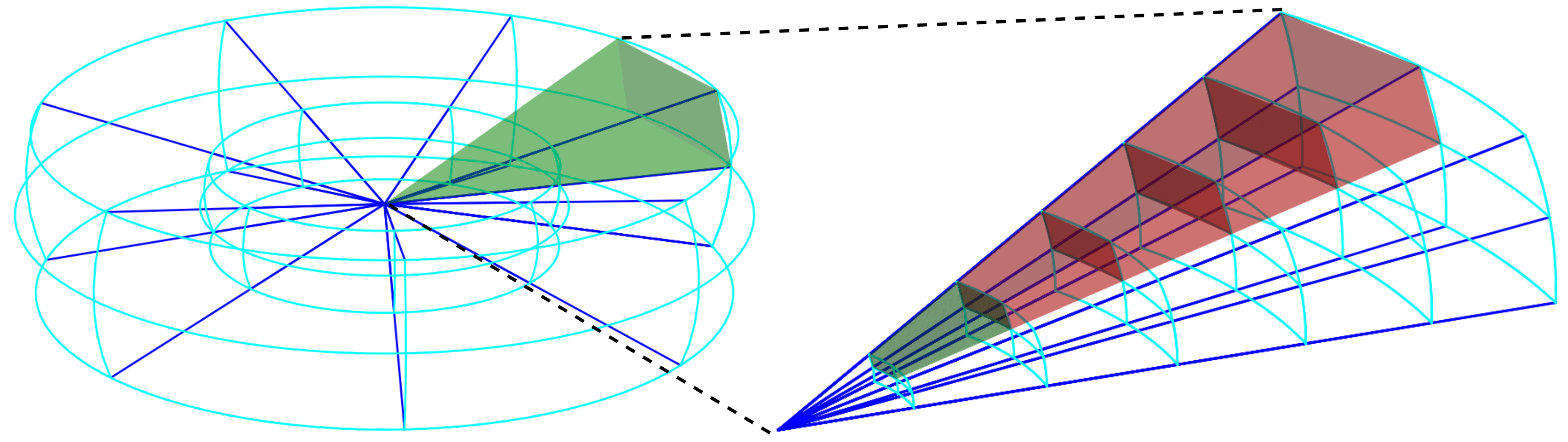}
    \caption{\textbf{(Left)} Spherical voxelization discretizes the space based on radius $r$ (distance from the origin), azimuth $\theta$ (horizontal angle), and elevation $\phi$ (vertical angle). 
    \textbf{(Right)} Occlusion handling in spherical representation is straightforward. If a voxel at coordinate \( (r, \theta, \phi) \) is occupied \textcolor{ForestGreen}{(Green)}, all voxels with the same azimuth and elevation but a larger radius (\( (r', \theta, \phi) \) where \( r' > r \)) will be occluded \textcolor{red}{(Red)}.
}
    \label{fig:spherical}
    \vspace{-10pt}
\end{figure} 

Our scene editing framework $\lidaredit$ is illustrated in Fig.~\ref{fig:method_overview}. We first introduce the spherical voxel representation (Sec.~\ref{sec:scene_representation}), which forms the foundation of our pipeline for efficiently modeling LiDAR projective geometry. Then, we present the three core components of $\lidaredit$. The first removes unwanted objects from the LiDAR scene (Sec.~\ref{sec:object_removal}) by identifying voxels occupied by the object and voxels occluded by them. We mask these voxels and use a generative inpainting model to fill the background.  
% Given an object to be removed, we identify the voxels occupied by the object's point cloud. We mask these voxels and all voxels occluded by them with an inpainting mask. We remove points in the masked voxels and use a generative point cloud inpainting model to fill the background. 

% Starting with the anchor point cloud $\gtpcloud$, we convert it into a spherical voxel representation (Sec.~\ref{sec:scene_representation}), which allows for effective handling of occlusion. 

% predict their occupancy. We train our model to only inpaint background scenes in this stage, so that the gaps left by removed objects are filled in by natural-looking background points. 
% We then identify the vehicle points to be removed and map them to their corresponding vehicle voxels. For each voxel, all points within it are replaced with a single background point, such as streets, walls, or trees. 
% This replacement process ensures that, instead of leaving unrealistic gaps, the removed objects are replaced with realistic background points. 
% However, due to the occlusion properties of LiDAR, not only the voxels containing vehicle points should be inpainted, but also those originally occluded by these points. We identify the full set of inpainted voxels through a process called De-occlusion Masking (Sec.~\ref{sec:de_occlusion}). 

In the second stage, we build an object library of dense full-shape point clouds (Sec.~\ref{sec:pc_completion}) for insertion with flexible poses. We extract partial-view point clouds using segmentation masks and transform them into full point clouds via learning-based completion.

% In the second stage, we collect an object library composed of dense full-shape point clouds for every object in all LiDAR scans (Sec.~\ref{sec:pc_completion}), so that we have a wide range of objects to be inserted back into the LiDAR scans with flexible poses. 
% % large dataset of full-view LiDAR point clouds of vehicles (Sec.~\ref{sec:pc_completion}). 
% % This is important because, to position any vehicle in any poses within a LiDAR point cloud scene, we need a diverse set of point clouds representing many vehicle types.
% We use the segmentation masks within each LiDAR scene to extract partial-view point clouds of vehicles. These partial views are then transformed into full point clouds using a learning-based point cloud completion method.

In the third stage, we insert objects into the LiDAR point cloud (Sec.~\ref{sec:object_insertion}) by selecting an object from our library and specifying a desired pose. The object is positioned on the ground, with resampling and occlusion handled through spherical voxelization.

% In the third stage, we insert new objects into the LiDAR point cloud (Sec.~\ref{sec:object_insertion}). We start by selecting an object from the full-view object library, and specifying a desired pose for its placement. The dense point cloud of this object is then positioned on the ground at the specified pose. The resampling of the inserted objects and the occlusion are then both handled in the spherical voxel representation. 
In our framework, moving an object is implemented as removing and re-adding it, which resolves background gaps and enables significant pose changes.

% It is worth noting that in our framework, moving an object to a different pose is treated as both removing and re-adding the same object. This approach helps resolve any gaps in the background and allows for significant pose changes.
% It is worth noting that the movement of an object to a different pose is also treated as removing and adding the same object in our framework. In this way, we can resolve the background gap and allow substantial pose changes. 
% Next, we identify the set of voxels that will be occluded by this newly placed object
% using a process called Occlusion Masking (Sec.~\ref{sec:occlusion}). Finally, we remove the occluded points from the scene using spherical projection.
% and remove the occluded points from the scene. 

\subsection{Spherical Voxelization} \label{sec:scene_representation}
The spherical voxel representation is core to our pipeline for efficient occlusion handling and density control (Fig.~\ref{fig:spherical}). 
% We represent each LiDAR point cloud as a set of discrete voxels using a spherical voxelization technique, as illustrated in Fig.~\ref{fig:spherical}. 
% Unlike traditional voxelization, which divides space into cubic or rectangular grids in Cartesian coordinates, 
Spherical voxelization discretizes space around the origin using three parameters: radius $r$, azimuth $\theta$, and elevation $\phi$. 
This mirrors LiDAR operation, where rays emit from the sensor at specific angles, allowing point resampling with a pattern consistent with real LiDAR sensors. Voxels sharing azimuth and elevation values form a single LiDAR ray, embedding occlusion relationships directly (Fig.~\ref{fig:spherical}-right). 
% Other common voxelization choices, like cubic or cylindrical partitions, do not have this property. 
While range-view representation could model occlusion, it loses 3D structure in 2D projection. Preserving this 3D structure enables effective feature learning and point cloud manipulation.  
% We deliberately choose the spherical representation over others, such as equirectangular, bird's-eye, or cylindrical representations. Equirectangular and bird's-eye views do not operate in 3D, leading to information loss. Cylindrical representation, while 3D, presents challenges in handling occlusion.

% The spherical representation, however, works well with both 3D data and occlusion handling due to its natural alignment with LiDAR scans physics. It mimics how a physical LiDAR ray would stop when hitting an object. As illustrated in Fig.~\ref{fig }, if a voxel at coordinates \( (r, \theta, \phi) \) is occupied, all voxels with the same azimuth and elevation but a larger radius, \( (r', \theta, \phi) \) where \( r' > r \), will be occluded. This property is highly useful in both the object removal and object insertion stages.

% \subsubsection{De-occlusion masking}\label{sec:de_occlusion}
% De-occlusion masking occurs during the object removal stage. Whenever we want to inpaint a vehicle voxel with a background point, we must also inpaint the other voxels that are occluded by the vehicle voxel (currently empty). This is necessary because those occluded voxels could have been occupied if the vehicle voxel were empty.

% \subsubsection{Occlusion masking}\label{sec:occlusion}
% Similarly, during object insertion, we need to apply occlusion masking to the voxels that will be occluded once the new object is placed in the scene. All points within these occluded voxels will be removed.

\begin{figure*}
    \centering
    \includegraphics[width=\textwidth]{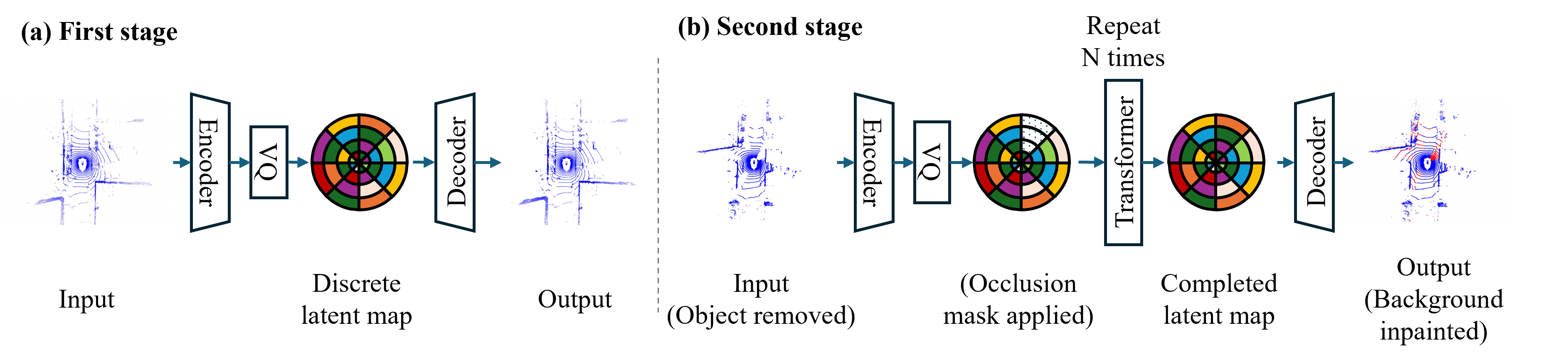}
    \caption{Overview of the background inpainting model. There are two stages in the training. (a) Use a VQ-VAE model to learn a discrete latent map in the bird's eye view. The colors represent the discrete latent codes. (b) Learn a multi-step autoregressive generation model that fills the masked tokens in the latent map and decode it to a full point cloud. Inpainted points are marked red. }
    \label{fig:inpaint_model}
    \vspace{-10pt}
\end{figure*}

\subsection{Object removal} \label{sec:object_removal}
% The object removal has two steps: identifying the area affected by the removal of an object and generating background points in this area that are consistent with the surrounding environment. We call these two steps \textit{de-occlusion masking} and \textit{background inpainting}, respectively. 
The object removal stage involves two steps: identifying the area affected by object removal and generating consistent background points in this area. We term these steps \textit{de-occlusion masking} and \textit{background inpainting}. 

\subsubsection{De-occlusion masking}\label{sec:deocclusion}
% Given the spherical voxelization of the input, a de-occlusion mask refers to a binary mask over the voxels, where the voxels occupied or occluded by the object to be removed are labeled positive. The occupancy of the masked voxels needs to be predicted because of the scene change (object removed) and the visibility change (occlusion removed). We first mask the voxels occupied by the object point cloud. Then, all voxels sharing the same azimuth $\theta$ and elevation $\phi$ indices with larger radius $r$ indices than the occupied voxels are also masked (see Fig.~\ref{fig:spherical}-right). 

% Given the spherical voxelization of a point cloud and the object to be removed, the de-occlusion mask refers to a binary mask highlighting the voxels whose occupancy needs to be predicted after the object is removed, i.e., the voxels occupied or occluded by the object. We first mask the voxels occupied by the object point cloud. Then, all voxels sharing the same azimuth $\theta$ and elevation $\phi$ indices with larger radius $r$ indices than the occupied voxels are also masked (see Fig.~\ref{fig:spherical}-right). 

Given spherical voxelization and the object to be removed, the de-occlusion mask highlights voxels whose occupancy needs prediction after removal—specifically, voxels occupied or occluded by the object. We first mask voxels occupied by the object, then mask all voxels sharing the same azimuth $\theta$ and elevation $\phi$ with larger radius $r$ indices than the occupied voxels (see Fig.~\ref{fig:spherical}-right). 

\subsubsection{Background inpainting}\label{sec:inpaint}
We approach background inpainting as a conditional generation problem using a method \baot{inspired by} the UltraLidar framework \cite{xiong2023ultralidar}, which is a LiDAR adaptation of MaskGIT\cite{chang2022maskgit} that employs transformer-based multi-step autoregression in discrete latent space. The training has two stages, as depicted in Fig. \ref{fig:inpaint_model}. First, we train a VQ-VAE \cite{van2017vqvae} to encode raw point clouds into a discrete latent feature map.  We represent this using a bird's-eye-view (BEV) 2D map with azimuth-radius coordinates, compressing elevation-angle-axis voxels to single latent points. This differs from physical top-down projection and UltraLidar's Cartesian voxelization. We use the original point cloud scans to train this encoder. 

% The second step is to learn a generative model in the latent space for background inpainting. Similar to MaskGIT \cite{chang2022maskgit}, a bi-directional transformer is trained to map an incomplete BEV latent feature map to a complete latent feature map, which can be decoded to a point cloud. The feature prediction is cast as a classification problem because the feature space is discrete. The prediction is iterative. We use a mask to keep track of which tokens in the BEV feature map need to be predicted. In every iteration, the transformer predicts the masked tokens from the known ones, but only a subset of the predicted tokens with the top classification confidence is preserved, and they will be treated as known tokens in the next iteration. The other tokens remain masked and need to be predicted in the following iterations until the feature map is complete. The BEV mask is initialized from the voxel mask that is used to remove points from the input point cloud. In training, we apply the voxel mask on random object-free voxels, so that the transformer is trained to predict \textit{only the background} for the masked tokens. At inference, we use the de-occlusion voxel mask introduced in Sec. \ref{sec:deocclusion}. 

The second step learns a generative model in latent space for background inpainting. Following MaskGIT~\cite{chang2022maskgit}, we train a bi-directional transformer to map incomplete BEV feature maps to complete ones, which decode to point clouds. Feature prediction is treated as a classification problem of the discrete latent codes with iterative processing. A mask tracks which tokens need prediction, with each iteration preserving only sampled high-confidence predictions as known tokens for subsequent iterations. The BEV mask initializes from the voxel mask used to remove points. During training, we apply the mask to random object-free voxels, training the transformer to predict only background for masked tokens. At inference, we use the de-occlusion mask from Sec.~\ref{sec:deocclusion}. 

% During training, we apply random masks among the object-free voxels so that the transformer is trained to predict \textit{only the background} for the masked tokens. In our framework, we take the point cloud with object points removed as input. In the encoded feature map, we mask out the BEV tokens corresponding to the masked voxels. Then, we run through the transformer to iteratively predict the masked tokens and decode the feature map to a full point cloud. During inference, we use the de-occlusion voxel mask to mask out. 

% While we follow the UltraLidar framework \cite{xiong2023ultralidar}, the object removal in UltraLidar is realized by replacing the token at the object position with a background discrete feature, which needs to be manually specified and does not explicitly inpaint the areas occluded by the removed objects. Our object-removal module resolves both problems. 
\bao{$\lidaredit$ inherits UltraLidar's two-stage framework \cite{xiong2023ultralidar}, while tailoring the system to generate realistic backgrounds after object removal by locating occluded areas via spherical voxelization and using background-only training. In comparison, UltraLidar simply replaces object tokens with manually-specified discrete background features without guaranteeing complete background inpainting. }

% $\lidaredit$ inherits UltraLidar's two-stage framework, while the spherical voxelization and background-only training strategy are designed explicitly for background inpainting after object removal. 

% 1. Background inpainting. Train UltraLidar to specialize in this task:
% UltraLidar: naively swap object code with background code, with no notion of occlusion. Just let the decoder do the job with no guarantee of occlusion

% Our work: explicitly constrain the occlusion by using spherical coordinate. Reason about what voxels are affected and inpaint all of them (with background only).

% --> We're more realistic.

\subsection{Object library construction}\label{sec:pc_completion}
We create a library of objects with full 3D shapes to be inserted at arbitrary poses in a LiDAR scene. The objects are collected from all LiDAR scans in our training data. We choose the point cloud as the shape representation of objects for two reasons. First, we can resample points from the shape conveniently through spherical voxelization without ray-casting on meshes or volumetric rendering on implicit fields. Second, point cloud completion techniques are well-established to facilitate full-shape generation. We apply a pretrained model of the state-of-the-art point cloud completion network, AnchorFormer \cite{chen2023anchorformer}, in our framework. 

\subsection{Object insertion}\label{sec:object_insertion}
% The insertion process consists of three main steps. First, we construct a dataset of full-view vehicle point clouds, using a point cloud completion neural network. Next, we specify the desired pose and place the vehicle point cloud into the LiDAR scene, enforcing it to be positioned on the ground. Finally, we filter out points that are occluded by the vehicle, including self-occlusion.

The insertion process has two major steps. First, given the desired pose to insert an object, we determine the height to make sure the object is placed on the ground. Second, we resolve the occlusion caused by object insertion, and resample the point cloud on the spherical voxel grids to yield a pattern consistent with the LiDAR scan. 
% \subsubsection{Vehicle point cloud completion} \label{sec:pc_completion}
% To freely place any vehicle at an arbitrary pose within a LiDAR point cloud scene, we need a diverse dataset of full-view point clouds representing a wide variety of vehicles. 
% We achieve this by first collecting partial-view vehicle point clouds from the nuScenes dataset~\cite{undefined}. This process involves loading the entire LiDAR point cloud and using instance segmentation masks to extract the points corresponding to each vehicle. Next, we use a learning-based point cloud completion method to convert each of these partial point clouds into full-view.

% Point cloud completion refers to the process of filling in the missing parts of a 3D object based on its partial observation: $\hat{\mathcal{P}} = g(\mathcal{P}_{\text{partial}}, \theta)$. Here, $\hat{\mathcal{P}}$ represents the predicted full point cloud, $\mathcal{P}_{\text{partial}}$ is the partial observation, and $g(., \theta)$ is the neural network responsible for the completion task.
% For our work, we leverage the state-of-the-art AnchorFormer~\cite{undefined} as the point cloud completion network. 
% % AnchorFormer employs discriminative nodes, known as anchors, to capture fine regional details of objects. Unlike methods that depend on a global feature vector, AnchorFormer dynamically places these anchors at both observed and unobserved points, refining them into a complete and detailed 3D structure.

\subsubsection{Object placement}\label{sec:object_placement}
% -- Ground detection and Height determination}
We assume that a user only specifies the x-y coordinate and the yaw angle of the desired insertion pose, and the insertion algorithm needs to figure out the rest of the degrees of freedom based on the existing background point cloud. We further assume zero pitch and roll angles, as commonly assumed in self-driving datasets \cite{caesar2020nuscenes, sun2020scalability}. Therefore, we are left with height to be determined. 

To achieve this, we use the segmentation mask to find the ground points. 
Given a desired object position $(x, y)$, we locate the nearest ground point in the x-y plane and align its $z$ value with that of the lowest point in the inserted object point cloud, allowing us to fully define the vehicle's pose in the 3D space. Finally, we place the completed full point cloud of the target object into the LiDAR scan. 

% To insert a new object into the LiDAR scene, we sample a full point cloud from the dataset and place it at the desired coordinates. However, to ensure realism, the vehicle must be placed on the ground (minimum $z$ value), not floating in midair.

% To achieve this, we first use the segmentation mask to detect ground points, which are then projected into a bird's-eye view (2D plane). We discretize these ground points into 2D grid cells, using a method similar to spherical voxelization, but without considering the elevation dimension.

% Given a desired vehicle coordinate $(x, y)$, we locate the nearest grid cell to this coordinate and select the point with the lowest z value within that grid cell. This z value is then used as the third dimension, allowing us to fully define the vehicle's placement at the $(x, y, z)$ coordinate.

\subsubsection{Resampling and resolving occlusion}
The point cloud, in its current state, presents two critical problems. First, the dense inserted point cloud does not match the surrounding environment in terms of scan pattern and density. Second, the occlusion relationships are incorrect.

These two problems can be efficiently fixed with our spherical voxelization technique. We locate the voxels occupied by the point cloud of the inserted vehicle. Then, we pick the center of each occupied voxel as the resampled points for the shape. Now, the inserted object has a pattern and density that are consistent with the overall LiDAR scan. In the end, we resolve all occlusions \baot{using spherical projection:} for each $(\theta, \phi)$ coordinate with at least one occupied voxel, an occlusion mask is applied to all voxels with radius $r>r_{min}$, where $r_{min}$ is the smallest radius of the occupied voxels in the $(\theta, \phi)$ 1D ray. All the masked voxels are set to empty. 

% After placing the full point cloud into the scene, we execute the occlusion masking procedure (explained in Sec.~\ref{sec:occlusion}) to identify the voxels that are occluded by the new vehicle. Once all the points within these voxels are removed, the remaining data represents a realistic synthetic LiDAR point cloud scene.

%% file: experiment.tex
% This paper aims at a novel LiDAR editing task that is different from the literature.
This paper introduces a novel LiDAR editing approach distinct from all existing methods.
Therefore, it is not straightforward to use a well-established evaluation protocol or compare $\lidaredit$ with previous works. We divide the experiments into two major components, object removal and object insertion, showing our framework generates realistic LiDAR scenes.
% , so that we can validate the effectiveness of each step and show that the overall framework provides a practical solution for LiDAR scene editing. 

% We conduct experiments on the nuScenes-LidarSeg~\cite{caesar2020nuscenes} dataset, which is composed of 40,000 point clouds sampled from 1,000 driving scenes. The point clouds are collected by a 32-beam Velodyne HDL32E LiDAR. We use the official training and validation splits. All results are trained on the LiDAR scans in the training set and evaluated using scans in the validation set. We focus on the manipulation of vehicle-type objects, including cars, trucks, and buses. The specific use of the dataset in each experiment is explained in the regarding sections. \bao{In all experiments, we use a voxelization resolution of xxx, xxx, and xxx.} 

\bao{We conduct experiments on the nuScenes-LidarSeg~\cite{caesar2020nuscenes} dataset, consisting of 40,000 point clouds from 1,000 driving scenes, captured by a 32-beam Velodyne HDL32E LiDAR. Using the official training and validation splits, all models are trained on the training set of size 28,130, and evaluated on the validation set of size 6,019. The dataset's specific use in each experiment is detailed in the relevant sections. }
% In all experiments, we use a voxel size of \shinghei{0.1m $\times$ 0.703 degrees $\times$ 1.29 degrees within the bounds $[0, 50]\times[0,360]\times[79.3, 121]$ along the radius, azimuth and elevation axes respectively}}. 
\mhz{In all experiments, we use spherical voxelization of resolution $512\times 512 \times 32$ within the range of $[0, 50]m\times[0,360]^\circ\times[79.3, 121]^\circ$ along the radius, azimuth and elevation axes respectively. }

\begin{figure}
    \centering
    \includegraphics[width=\linewidth]{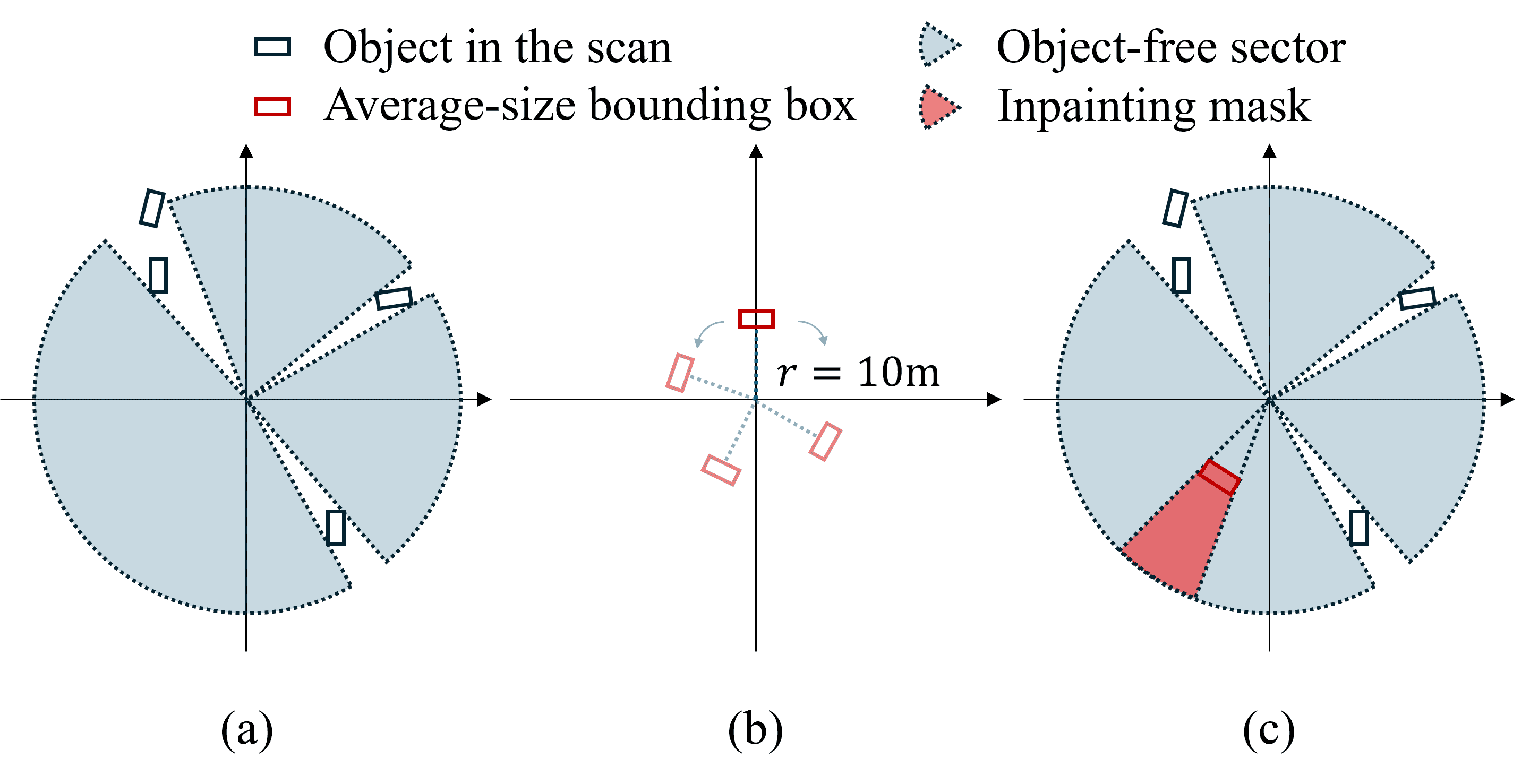}
    \caption{Illustration of the inpainting mask creation process in the background inpainting experiment. (a) shows the object-free sectors in the original scan. (b) shows that we use a nominal bounding box of the average size at 10 meters distance to create the mask. The bounding box can be rotated to fit in an object-free sector. (c) shows an example inpainting mask created from a rotated bounding box. }
    \label{fig:inpaint_mask}
    \vspace{-7pt}
\end{figure}

\begin{figure}
    \centering
    \includegraphics[width=\linewidth]{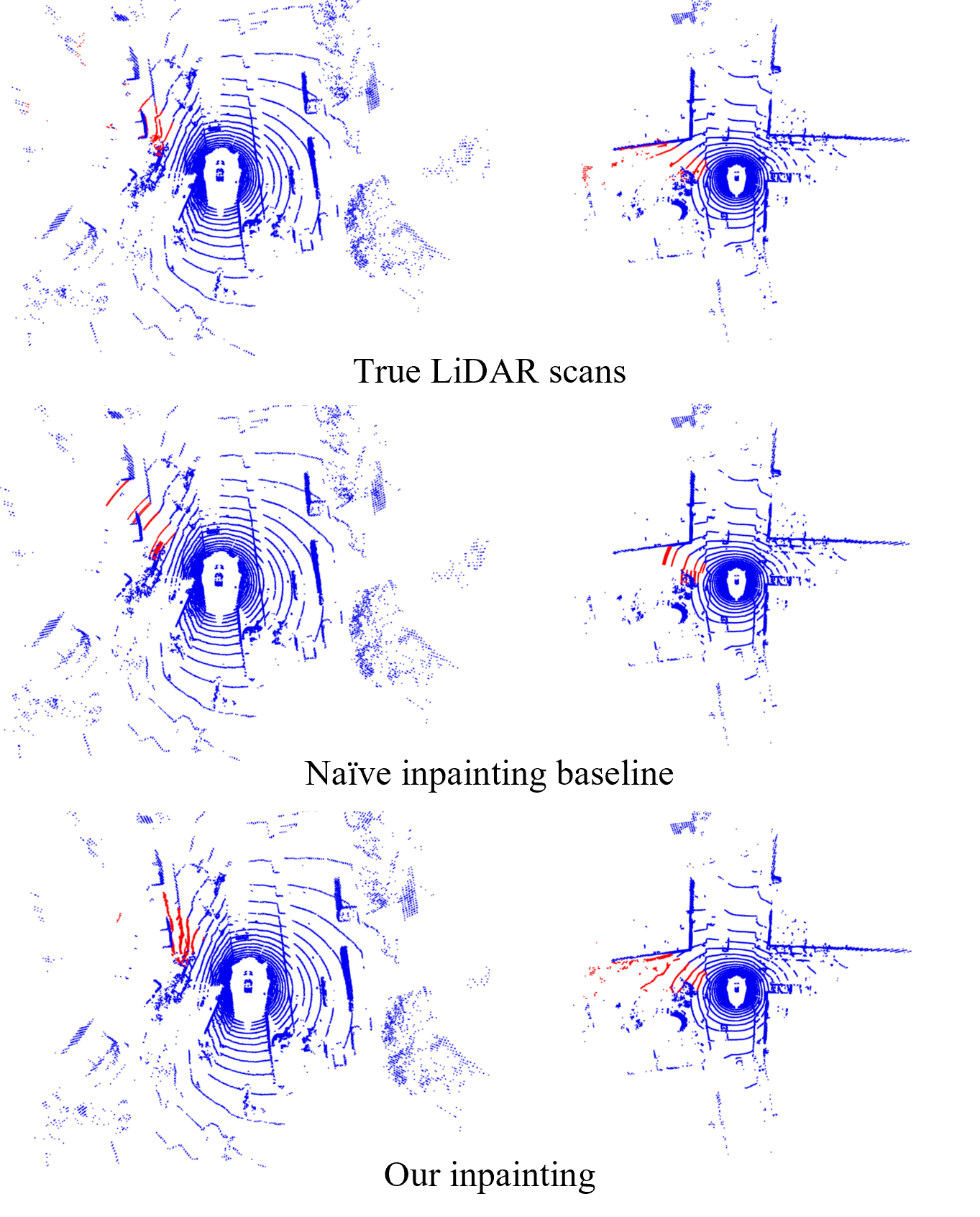}
    \caption{Qualitative results of background inpainting on nuScenes dataset. Red highlights the points in the inpainting mask. Our prediction fits naturally with the surrounding environment and is close to the actual scans. }
    \label{fig:inpaint}
    \vspace{-7pt}
\end{figure}

\subsection{Object removal and background inpainting}
% The goal of this experiment is to confirm that $\lidaredit$ can inpaint realistic background points after the vehicles are removed. 
% The tricky thing is that the ground truth background behind a foreground object does not exist. Therefore, we create artificial de-occlusion masks over the background area in a LiDAR scan, and ask our model to reconstruct the masked background points. In this way, we have the ground-truth background for meaningful evaluation metrics. 
We design two experiments to evaluate the quality of our inpainted background and to verify that our pipeline can generate object-free point clouds after object removal and background inpainting, respectively. 

The first experiment evaluates $\lidaredit$'s ability to inpaint realistic background after vehicle removal. Since ground truth background behind real objects is not available, we create artificial de-occlusion masks over background areas and task our model with reconstructing these masked points, giving us ground-truth for evaluation. 

To create masks resembling real object occlusions, we use bounding boxes from nuScenes objects (Fig. \ref{fig:inpaint_mask}). We calculate the average size of all ground truth vehicle bounding boxes, then place a box of the average size at a fixed 10-meter distance with heading orthogonal to viewing direction to maximize azimuth occlusion. We rotate this box until it falls within an object-free sector of the LiDAR scan.

% In order to make the artificial masks similar to the ones caused by the occlusion of real objects, we create the masks from the bounding boxes of objects in nuScenes. 
% This process is illustrated in Fig. \ref{fig:inpaint_mask}. We calculate the average size of the ground truth 3D bounding boxes of all vehicles in nuScenes. Then, we put the bounding box at a fixed distance, with the heading orthogonal to the viewing direction, so that the box occludes the most azimuth range. We rotate the bounding box around the ego vehicle till the bounding box falls into an object-free sector of the LiDAR scan to be masked. We pick the distance as 10 meters to create a substantial occlusion. 

% Specifically, we collect the 3D bounding boxes of all objects from all LiDAR scans into an object pool. Then we sample one such bounding box and put it into a LiDAR scan to create an occlusion mask. To guarantee that the masked area is object-free, we need to make sure the size of the azimuth angular interval occupied by the sampled bounding box is smaller than that of the largest object-free sector in the LiDAR scan. Then, we rotate the sampled bounding box around the ego vehicle till the bounding box falls into an object-free sector. In this way, the occlusion mask will be of the same size as caused by the real bounding box and in the object-free area. To create substantial occlusions, we only sample bounding boxes within a certain distance to the ego vehicle. We pick the threshold as 10 meters in this experiment.  

\bao{We evaluate our generated background against ground truth using both statistical and perceptual metrics. Statistical metrics quantifythe global structural differences between the two sets of background poins, while perceptual metrics assess differences in the latent feature space, capturing both global and local structures.}
% For the statistical metrics, we leverage the Maximum-Mean Discrepancy (MMD) and the Jensen-Shannon Divergence (JSD). Similar to \cite{xiong2023ultralidar}, we construct a 2D histogram along the ground plane and use the voxel occupancy instead of the number of points for the bin count. However, our evaluation has two differences from that in \cite{xiong2023ultralidar}. First, our 2D histogram is in the azimuth-radius coordinate. Second, we manually set the count for all bins that are not in the generated area to zero and re-normalize the histogram to sum to one, because we only generate the point cloud for a masked area instead of generating the whole point cloud from scratch. For the perceptual metrics, we use the Frechet Sparse Volume Distance (FSVD) and the Frechet Point-based Volume Distance (FPVD). 
\bao{For statistical metrics, we use Maximum-Mean Discrepancy (MMD) and Jensen-Shannon Divergence (JSD). Following \cite{xiong2023ultralidar}, we construct a 2D histogram along the ground plane, using voxel occupancy instead of point count for bin values. Our evaluation differs in two ways: (1) our 2D histogram is in azimuth-radius coordinates, and (2) we zero-out bins outside the generated area and re-normalize the histogram, as we generate only a masked region rather than an entire point cloud. For perceptual metrics, we use Frechet Sparse Volume Distance (FSVD) and Frechet Point-based Volume Distance (FPVD).}
Following \cite{ran2024lidardiffusion}, FSVD is evaluated using MinkowskiNet \cite{choy2019minkowski}, and FPVD is evaluated using SPVCNN \cite{tang2020searching}, both pretrained on nuScenes.

\begin{table}[t]
    \centering
    \resizebox{0.9\columnwidth}{!}{
    \begin{tabular}{lcccc}
    \toprule
         & \multicolumn{2}{c}{Perceptual} & \multicolumn{2}{c}{Statistical} \\
    Metrics  & FSVD $\downarrow$           & FPVD $\downarrow$        & JSD $\downarrow$         & MMD $\downarrow$       \\ \midrule
    Baseline & 0.189           & 0.191          &   0.615        &         6.85$\times 10^{-6}$   \\
    Ours     & \textbf{0.169}           & \textbf{0.173}          &   \textbf{0.584}        &         \textbf{6.61}$\mathbf{\times 10^{-6}}$      \\
    \bottomrule
    \end{tabular}    }
    \caption{Quantitative evaluation of the background inpainting task on nuScenes dataset. $\downarrow$ means the lower the better. }
    \label{tab:obj_removal}
    \vspace{-10pt}
\end{table}

\begin{table}[]
    \centering
    \resizebox{\columnwidth}{!}{
    \begin{tabular}{lcc}
    \toprule
    Data generation method                                 & Ours  & UltraLidar~\cite{xiong2023ultralidar} \\ \midrule
    Number of objects detected per frame &  2.6       &  15.5   \\
    \bottomrule
    \end{tabular}}
    \caption{\baot{Quantitative results of object removal of our method evaluated using a pretrained object detector, VoxelNext~\cite{chen2023voxelnext}. The unconditionally generated point clouds from UltraLidar are compared as a reference.}}
    % \caption{The number of objects detected by a pretrained VoxelNext~\cite{chen2023voxelnext} in our object-free point clouds after background removal and background inpainting, compared with unconditionally generated point clouds from UltraLidar. }
    \label{tab:obj_empty_det}
    \vspace{-10pt}
\end{table}

\bao{As no prior work addresses this exact task, we compare our method to a naive inpainting baseline, which copies and tiles a neighboring object-free point cloud sector into the masked area.
Using the nuScenes validation set for evaluation, Tab.~\ref{tab:obj_removal} shows that our method outperforms the baseline across all metrics, demonstrating its effectiveness. Fig. \ref{fig:inpaint} provides qualitative comparisons, highlighting how our model inpaints the background based on the context, producing more natural and realistic point clouds.}

In the second experiment, we use de-occlusion masks from ground-truth object bounding boxes to generate object-free point clouds. We then use an object detection model, VoxelNext~\cite{chen2023voxelnext}, to count objects in the point clouds, and the results are in Tab.~\ref{tab:obj_empty_det}. 
% Ideally, the number should be zero, but a small non-zero number is reported due to false positives, which is expected. 
Ideally, the count should be zero; however, a small nonzero value is observed due to false positives, which is expected.
\baot{For reference, UltraLidar’s unconditionally generated point clouds, which are expected to contain objects, yield an average of 15.5 detected objects.} % Minghan: we are trying to beat UltraLidar here. 

% As a reference, the point clouds generated by UltraLidar unconditionally, which are expected to contain objects, have 15.5 detected objects on average.

\begin{figure}
    \centering
    \includegraphics[width=\linewidth]{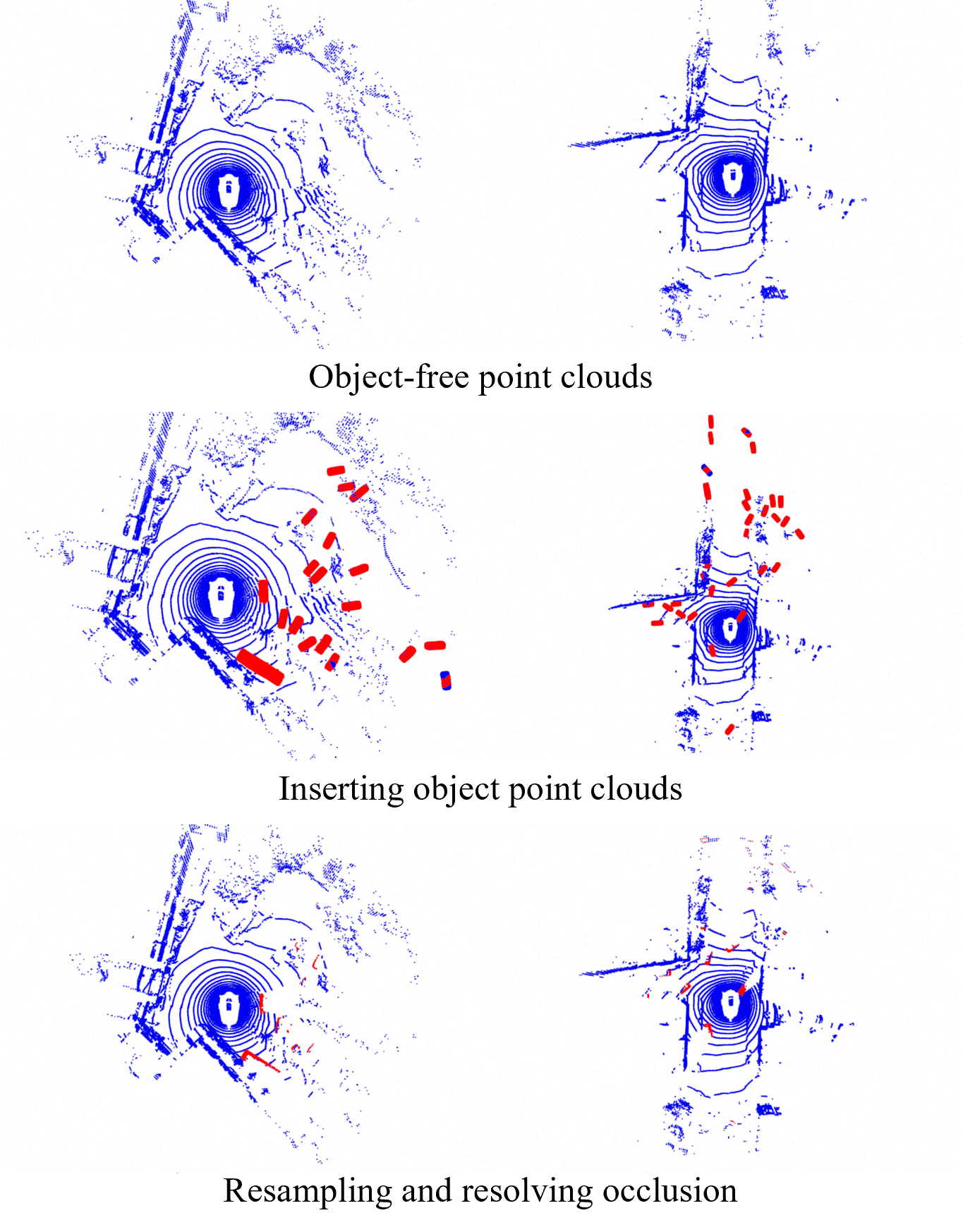}
    \caption{Object insertion qualitative results. Inserted points are highlighted in red. }
    \label{fig:insertion_qualitative}
    \vspace{-5pt}
\end{figure}

\begin{table}[t]
    \centering
    \resizebox{\columnwidth}{!}{
    \begin{tabular}{llccc}
    \toprule
    Training data      & Testing data           & mAP $\uparrow$  & car $\uparrow$ & non-car $\uparrow$ \\ \midrule
    nuScenes  & Ours & 0.340 & \textbf{0.785} & 0.290 \\
    nuScenes  & nuScenes & 0.361 & 0.672 & 0.326 \\
    Ours + nuScenes & nuScenes & \textbf{0.368} & 0.672 & \textbf{0.334} \\ \bottomrule
    \end{tabular}}
    \caption{Object detection performance (average precision) of VoxelNext~\cite{chen2023voxelnext} trained and tested on different datasets. }
    \label{tab:obj_insert}\vspace{-10pt}
\end{table}

\subsection{Object insertion}
The second part of our framework is object insertion. It is designed to allow users to freely specify the number, type, and pose of the objects to be inserted. 
\mhz{In this experiment, we only manipulate the \textit{car} category because AnchorFormer's~\cite{chen2023anchorformer} point cloud completion is less reliable on other categories, but our framework is agnostic of the choice of completion models. }In Fig. \ref{fig:insertion_qualitative}, we show some qualitative examples of object insertion. 

The choice of object placement will also affect the quality of the generated data and matters if we want to evaluate the generated data quantitatively. Therefore, we design different object insertion strategies when we evaluate different aspects of the object insertion algorithm. Specifically, we have two experiments. The first reveals the domain gap between the generated synthetic data and the real data, and the second showcases the value of our synthetic data in improving autonomous driving downstream tasks such as object detection. 

\subsubsection{Domain gap analysis}
% The realism of the generated data depends on both the object insertion algorithm and the number and pose of the objects to be inserted. 
To rule out the impact of the choice of object placement on the realism of the generated data, we insert objects at the same pose as in the true LiDAR scans. Specifically, we first remove all objects and inpaint the background. Then, we sample random objects from the object library and insert them into the LiDAR scan at the pose of the removed objects. 
% In this way, we obtain a synthesized point cloud with the same pose layout of objects but with different objects. 
\baot{In this way, we obtain a synthesized point cloud that maintains the original pose layout of objects while replacing the objects with different ones. }
\baot{We apply this procedure to the entire nuScenes validation set, producing a \textit{synthetic validation set} of the same size as the original.}
% We apply this procedure to the nuScenes validation set and generate the same number of point clouds for comparison. }

% We evaluate the quality of our generated synethetic point clouds by running  
% VoxelNext \cite{chen2023voxelnext} pretrained on nuScenes. 
We assess the quality of our generated synthetic point clouds using \baot{the object detector} VoxelNext \cite{chen2023voxelnext}, pretrained on nuScenes. We evaluate VoxelNext on both the (1) real and (2) synthetic nuScenes validation sets.
\mhz{In the first two rows of Tab.~\ref{tab:obj_insert}, we can see that \baot{the mAP (mean average precision over all categories) remains similar when tested on the original validation set compared to our synthetic data, indicating a small domain gap}. The car category's higher AP shows that our inserted object point clouds have a small domain gap with the original ones, but they may also be too simple due to the lack of ray-drop modeling, indicating a direction for future improvements. The slightly lower APs of other categories are mostly due to the increased false positives caused by the background inpainting, which is helpful as we will see in Sec.~\ref{sec:downstream}. }

% We evaluate the quality of the generated point clouds by running an object detection network VoxelNext \cite{chen2023voxelnext} pretrained on nuScenes. The results are in Tab. \ref{tab:obj_insert_generalization}. As expected, the network performs differently on the true nuScenes dataset compared to our generated dataset. However, the results are comparable, with categories both outperforming and underperforming, suggesting a small domain gap between the two datasets.

% \subsubsection{\Value for downstream tasks}\label{sec:downstream}
\subsubsection{\baot{Values on autonomous driving downstream tasks}}\label{sec:downstream}
We aim to demonstrate the practical value of our generated data in improving perception algorithms. 
% \mhz{We use our generated dataset to pretrain VoxelNext for 15 epochs. We then train the model on the original nuScenes dataset for 40 epochs with and without the pretraining. }
% To do this, we combine the real-world dataset with our generated dataset to fine-tune a detection network and evaluate its performance. 
\baot{We compare two VoxelNext models, trained from scratch, using two different training procedures. The first model is trained \textit{only} on real nuScenes data for 40 epochs. The second model is first pretrained on our synthetic training dataset (details provided below)  for 15 epochs; then trained on real nuScenes data for an additional 40 epochs, following the same training schedule as the first model.}
In this experiment, we design our synthetic data to have different object layouts than the real data to introduce more variability. Specifically, we randomly perturb the target object poses by up to 2.5 meters and 45 degrees in yaw angle, while ensuring they remain in the ground area on the BEV map. The inserted objects are sampled from the object library. \mhz{We apply this procedure to the entire nuScenes training set, generating a \baot{\textit{synthetic training set} of the same size} for pretraining. Tab. \ref{tab:obj_insert} shows that pretraining on our generated data (third row) yields superior performance compared to training exclusively on real nuScenes (second row). This shows that our synthetic data provides significant value in building more powerful algorithms. Interestingly, the improvement mostly comes from non-car categories by allowing the model to better suppress false positives. The inserted car objects, which are slightly easier to detect than the real ones, did not change the performance of the final model, but we expect it to improve with more detailed LiDAR modeling like ray-drop.}

% \subsection{Ablation study}
% How to show spherical is better? 
% Statistic and perceptual metrics against cylinder view?

%% file: conclusions.tex
% In this paper, we contribute a novel framework for LiDAR scene editing, generating realistic synthetic data with novel object layouts in a real-world environment. 
% We leverage generative models to fill in the missing information about both the background and the objects, and adopt a spherical voxelization model to handle LiDAR projection geometry efficiently.  
% We demonstrate the effectiveness of our framework through experiments on a large-scale self-driving dataset, nuScenes, showing that the generated LiDAR point clouds closely resemble real-world data. We show incorporating these synthetic LiDAR scans as additional training data leads to significant improvements in object detection performance in real-world data.
% For future work, we plan to extend the object editing task beyond vehicle-type objects. We will also use novel view synthesis to build a flexible framework, where the novel-view information may come from temporarily continuous observations when available and from prior knowledge embedded in generative models otherwise.

% \bao{This paper contributes a novel framework for LiDAR scene editing, generating realistic synthetic data with novel object layouts in real-world environments. Using generative models, we fill in the missing information about both the background and the objects, and employ spherical voxelization for efficient LiDAR projection handling. Experiments show our generated synthetic LiDAR point clouds closely resemble real-world data while adding diversity.}
This paper introduces a novel paradigm for LiDAR data generation by editing the object layouts in real-world LiDAR scans. Following this paradigm, our $\lidaredit$ framework generates data with a small domain gap to the real world and gives users high controllability. Experiments show practical value of the generated data in downstream autonomous driving tasks. Incorporating ray-drop effects, optimizing object shape models, and automating the pose layout generation are interesting improving directions for future work. 
% , improving object detection performance when used as additional training data. 
% Future work includes extending object editing beyond vehicles and incorporating novel view synthesis for a more flexible framework using temporal observations or generative model priors.